\documentclass[10pt,twocolumn,letterpaper]{article}

\usepackage{cvpr}      
\definecolor{cvprblue}{rgb}{0.21,0.49,0.74}
\usepackage[pagebackref,breaklinks,colorlinks,allcolors=cvprblue]{hyperref}

\usepackage{caption}
\usepackage{cuted}
\usepackage{caption} 
\usepackage{float}

\usepackage{graphicx}
\usepackage{booktabs}
\usepackage{amsmath}
\usepackage{amssymb}
\usepackage{algorithm}
\usepackage{algpseudocode}
\usepackage{multirow}
\usepackage{xspace}
\usepackage{enumitem}
\usepackage[normalem]{ulem}
\usepackage{makecell}
\usepackage{adjustbox}
\usepackage{tabularx}
\usepackage{tcolorbox}
\usepackage{lipsum} 
\tcbuselibrary{skins}

\useunder{\uline}{\ul}{}

\def\paperID{} 
\def\confName{CVPR}
\def\confYear{2026}

\title{
InstantRetouch: Efficient and High-Fidelity Instruction-Guided Image Retouching with Bilateral Space
}

\author{Jiarui Wu\textsuperscript{1,2}, Yujin Wang\textsuperscript{1,\dag}, Ruikang Li\textsuperscript{1,2}, Fan Zhang\textsuperscript{1}, Mingde Yao\textsuperscript{2,3}, Tianfan Xue\textsuperscript{2,1,3} \vspace{+4pt} \\
\vspace{+4pt}
\textsuperscript{1}Shanghai AI Laboratory,\textsuperscript{2}CUHK MMLab,\textsuperscript{3}CPII under InnoHK\\
{\small \textsuperscript{\dag}Corresponding author.}
\vspace{-20pt}
}

\vspace{-20px}
\begin{document}

\twocolumn[{
\renewcommand\twocolumn[1][]{#1}
\maketitle
\begin{center}
    \centering
    \captionsetup{type=figure}
    \includegraphics[width=0.97\linewidth]{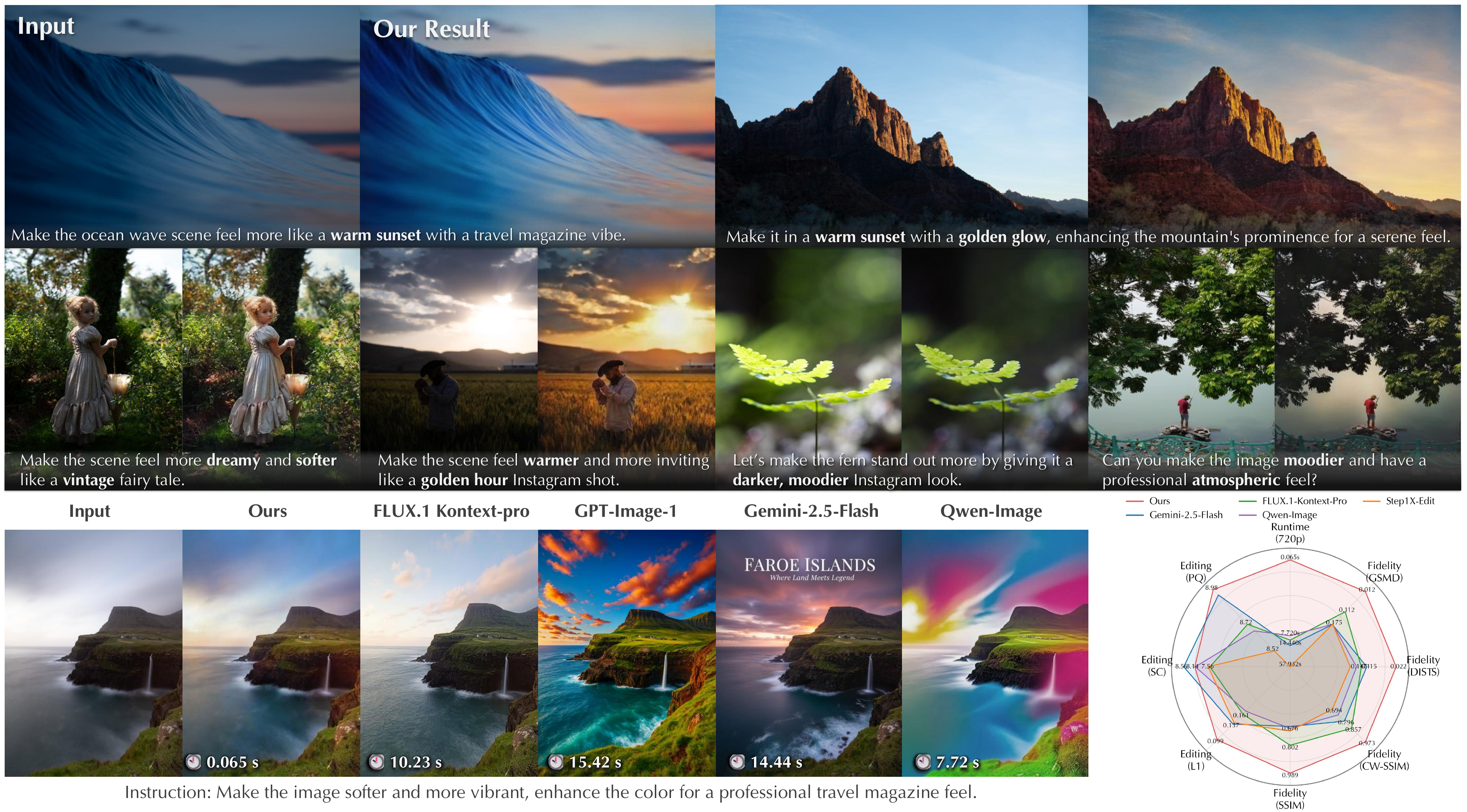}
    \vspace{-10pt}
    \captionof{figure}{ 
    Comparing our method with state-of-the-art image editing methods~\cite{labs2025flux, wu2025qwen, comanici2025gemini, liu2025step1x}. As shown in the upper part, our method follows text instructions to generate visually pleasing retouching results while preserving high fidelity, whether for natural landscapes or portraits. In contrast, as demonstrated in the lower section, other state-of-the-art methods modify the original content. Finally, as depicted in the multi-dimensional comparison chart, our method outperforms others in terms of \textbf{\textit{fidelity}}, \textbf{\textit{quality}}, and \textbf{\textit{speed}}.}
    \vspace{-1px}
    \label{fig:teaser}
\end{center}
}]

\vspace{-15pt}
\begin{abstract}
\vspace{-20pt}

Language-guided photo retouching aims to adjust color and tone while preserving geometry and texture. Recently, diffusion-based retouching shows a superior visual quality, but often struggles with both fidelity issues due to its generative nature and efficiency because of its iterative sampling process.
In this work, we propose an efficient and fidelity-preserving retouching method using bilateral space manipulation, which is both compact and content-decoupled. Specifically, instead of directly editing pixels or image latents, our model predicts a low-resolution bilateral grid of affine transforms, which are sliced using a learned guidance map and then applied to the full-resolution image. This approach yields both high fidelity and improved efficiency.
To retain strong priors of a pretrained generative model, we distill a multi-step diffusion model into our bilateral grid framework using Variational Score Distillation, complemented by a prompt alignment loss to guide instruction-following behavior. Additionally, we introduce a new benchmark and evaluate our method across multiple dimensions: fidelity, instruction following, and efficiency.
Compared to the latest editing methods, like Gemini-2.5-Flash (Nano-Banana), our method can avoid content drift, significantly improve latency, and generate visually pleasing edits, while maintaining a high level of fidelity. Project page: \url{https://openimaginglab.github.io/InstantRetouch/}.

\end{abstract}

\vspace{-10pt}
\section{Introduction}
\label{sec:intro}
\vspace{-5pt}

The ability to automatically retouch photos using natural language instructions represents a significant advancement over traditional image enhancement algorithms~\cite{zeng20203dlut, ouyang2023rsfnet}, which often lack expressive, fine-grained control. This paradigm shift has been driven by large-scale diffusion models~\cite{rombach2022high, brooks2023instructpix2pix}, capable of producing expressive and visually pleasing results guided by user instructions. Recent work continues to scale up these generative models for general-purpose image editing, as seen in Step-1X-Edit~\cite{liu2025step1x}, FLUX.1-Kontext~\cite{labs2025flux}, Qwen-Image~\cite{wu2025qwen}, or Gemini-2.5-Flash~\cite{comanici2025gemini}. These models exhibit a remarkable ability in general-purpose editing, such as adding or removing objects, often producing results that are indistinguishable from real images.

Still, these generative models exhibit limitations in fidelity and efficiency when applied to image retouching. First, for photo retouching, changes must be restricted to photometric adjustments without affecting geometry or texture. Existing generative editing models, however, may not adequately disentangle these edits, leading to unwanted content drift, as shown in~\cref{fig:teaser}. Second, these models, often based on iterative diffusion processes, are computationally expensive and slow, limiting their application for high-resolution image retouching.

These limitations arise because generative editing directly modifies the variational latent of the input image in the diffusion process~\cite{rombach2022high}. The latent representation consists of both actual image content and photometric information (brightness, color, etc.), which is unnecessarily large for retouching, slowing down the process. Manipulating in latent representation may also introduce the risk of changing actual visual content, texture, or geometric structure. Instead, retouch editing should only operate on a smaller representation that only focuses on visual appearance, without content information.

Therefore, we propose to only predict the parameters of a transformation in a compact and content-decoupled \emph{bilateral space}, instead of directly touch original image content. 
The bilateral manipulation space~\cite{chen2007real, chen2016bilateral, gharbi2017deep} is instantiated as a low-resolution 3D bilateral grid of affine transforms. A learned guidance map slices this grid to produce per-pixel affine coefficients that are applied to the full-resolution image, enabling complex tonal adjustments. 
This representation is exceptionally efficient even at 4K resolution and ensures high fidelity by design. As shown in Fig.~\ref{fig:teaser}, our solution based on bilateral grids is 70-800 times faster and better preserves fidelity compared with baselines.

While the bilateral space offers an ideal representation for retouching, generating visually pleasing results from instructions still requires the rich semantic priors from diffusion models. However, their slow, iterative inference is fundamentally at odds with our desired efficiency. We resolve this conflict by distilling a multi-step diffusion model into a fast, one-step generator that directly predicts the bilateral grid. In this way, we can leverage the \emph{rich diffusion priors} for visually pleasing results guided by instruction, along with the \emph{fidelity and efficiency} advantages of the bilateral space.

To enable this distillation, we first curate a large-scale, high-quality instruction-retouching dataset to fine-tune a multi-step teacher diffusion model. We then transfer its knowledge to an efficient student network which outputs the bilateral grid in a single forward pass, using a one-step bilateral distillation framework. In this one-step distillation, we employ Variational Score Distillation (VSD)~\cite{yin2024dmd, wang2023prolificdreamer} as the core objective, which we augment with a CLIP-based~\cite{radford2021learning} prompt alignment loss. This provides crucial semantic supervision to improve instruction following, particularly for ambiguous or stylistic prompts where pixel-level signals are weak. In addition, we design a bilateral loss to better regularize bilateral grid prediction. At last, we design a progressive distillation strategy to ensure training stability.

To evaluate performance on the instruction-guided retouching task, we introduce a new benchmark, iRetouch, composed of diverse real-world instruction-guided retouching scenarios. We assess models along three key axes: content fidelity, measured by the preservation of original texture and geometry; instruction following, evaluated via text-image alignment metrics and human preference studies; and efficiency, quantified by latency at various resolutions. As demonstrated in Fig.~\ref{fig:teaser}, our method is 70-800 times faster than large editing models~\cite{wu2025qwen,labs2025flux, liu2025step1x, comanici2025gemini, hurst2024gpt4o, brooks2023instructpix2pix} and achieves superior content fidelity, all while maintaining comparable instruction-following performance.

\vspace{-5px}
\section{Related Works}
\label{sec:related}
\vspace{-5px}
\noindent\textbf{Instruction-based image editing.}
Image editing enables intuitive image modifications driven by language. Early works, such as InstructPix2Pix~\cite{brooks2023instructpix2pix}, fine-tuned diffusion models by creating paired instruction-image datasets. Subsequent research~\cite{li2023moecontroller, mao2025ace++, guo2024focus, zhao2024ultraedit, duan2025diffretouch} primarily focused on architectural optimizations to improve control granularity and consistency, while others~\cite{geng2024instructdiffusion, sheynin2024emu, zhang2023magicbrush, chakrabarty-etal-2023-learning} concentrated on data-driven enhancements, expanding the range of instructions and diversifying editing examples. Additionally, some approaches~\cite{li2023instructany2pix, huang2024smartedit, zhang2025context, hu2025image} integrated large language model reasoning with diffusion-based image synthesis, while others leveraged chain-of-thought (CoT) reasoning~\cite{fu2023guiding} to improve the model's reasoning ability for handling more complex editing tasks.
Flow-edit~\cite{kulikov2025flowedit} constructs an ordinary differential equation to map the source and target distributions, reducing transport costs in text-driven editing. JarvisArt~\cite{lin2025jarvisart}, on the other hand, combines a multi-modal large language model (MLLM)-driven agent that understands user intent and intelligently coordinates over 200 retouching tools.
Recently, image editing has increasingly shifted toward large models with multi-modal fusion~\cite{zhang2025nexus, wang2025seededit, liu2025step1x, labs2025flux, bai2025qwen2, deng2025emerging, comanici2025gemini}. For example, FLUX.1 Kontext~\cite{labs2025flux}, as a generative flow matching model, integrates both image generation and editing tasks into a unified architecture, handling both local editing and generative in-context tasks.

\noindent\textbf{Image retouching.} 
Automating the complex task of image style adjustment has seen varied approaches. Early methods like 3D LUTs~\cite{zeng20203dlut, ouyang2023rsfnet} were fast but confined to fixed styles, while generative models~\cite{karras2019style} often lack sufficient interpretability and usually alter the original content of the image. More recent works utilized reinforcement learning to automate editing~\cite{hu2018exposure, wu2024goal, kosugi2020unpaired}. Tseng~\etal~\cite{tseng2022neural} used neural networks to proxy different image processing modules and optimized the image processing pipeline parameters using a style loss function.
However, those methods mentioned above typically handle a single style during training and cannot offer flexible control based on instructions.

\vspace{-5px}
\section{Method Overview}
\vspace{-5px}
\begin{figure*}[t]
\centering
  \includegraphics[width=1.0\textwidth]{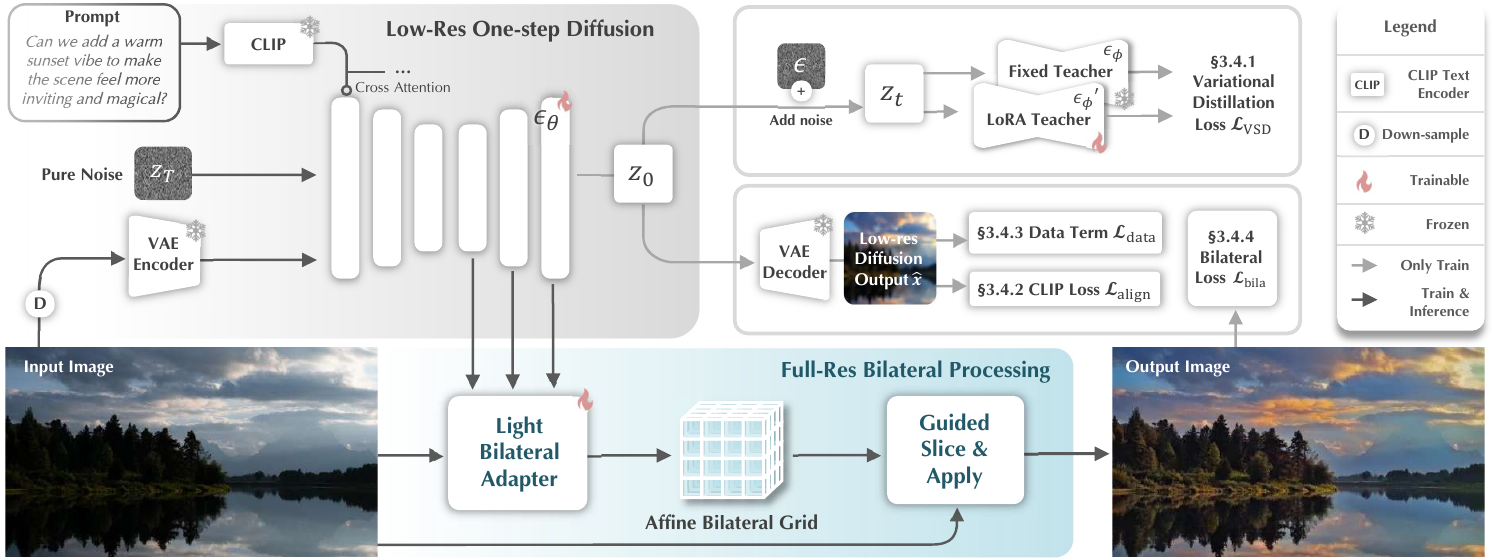}
  \vspace{-15pt}
  \caption{Our framework distills a multi-step diffusion teacher into a fast, one-step generator composed of two synergistic branches. The low-resolution diffusion branch processes the input image and text instruction to understand the edit, and then uses a light bilateral adapter to predict the parameters of a bilateral grid. The full-resolution branch then applies this grid to the original high-res image, producing the final high-fidelity result. We use Variational Score Distillation (VSD) to transfer the teacher's knowledge and a CLIP-based language alignment loss to ensure instruction alignment.}

  \vspace{-15pt}
  \label{fig:framework}
\end{figure*}

Our goal is to leverage the \emph{rich diffusion priors} for instruction-guided editing while retaining the \emph{fidelity and efficiency} of the bilateral space. To this end, our method distills a multi-step diffusion model into a fast, one-step generator, $G_\theta$, that directly predicts a bilateral grid. The process unfolds in two main stages. First, we curate a large-scale, high-quality instruction-retouching dataset (Sec.~\ref{sec:finetune}) to fine-tune a multi-step diffusion teacher, $\epsilon_{\phi}$ (Sec.~\ref{sec:method-teacher}). Second, we distill the knowledge from this teacher into our one-step bilateral grid generator (Sec.~\ref{sec:method-generator}) using a novel distillation framework (Sec.~\ref{sec:method-onestep}).

As illustrated in Fig.~\ref{fig:framework}, our generator $G_\theta$ consists of two synergistic branches: a low-resolution one-step diffusion branch for semantic understanding and retaining rich diffusion priors, and a full-resolution bilateral processing branch that applies the learned edit to deliver high-fidelity retouching on high-resolution input. However, directly training the proposed bilateral processing network may introduce instability in training; we instead adopt a progressive training strategy. We first train the low-resolution branch by minimizing the Variational Score Distillation loss (Sec.~\ref{sec:method-vsd}), a data loss (Sec.~\ref{sec:method-data}), and our prompt alignment loss (Sec.~\ref{sec:method-clip}). We then jointly optimize both branches, adding a bilateral loss (Sec.~\ref{sec:method-bila}) to optimize the full-res bilateral branch.

\vspace{-5px}
\subsection{Training Dataset}
\label{sec:finetune}
\vspace{-5px}

Training diffusion models relies on large-scale data. Existing instruction-editing datasets primarily focus on object-level or geometric edits and lack the fine-grained, high-fidelity examples needed for photo retouching. We therefore construct a new dataset of $\sim200$K triplets $(x, x^{\star}, c_T)$, where $x$ is the input image, $x^{\star}$ is a high-quality retouched target, and $c_T$ is a textual instruction. Our dataset is built via a controlled degradation process.

\noindent\textbf{High-quality targets.} We curate visually pleasing images from public datasets and the web, filtered by no-reference image quality metrics MUSIQ~\cite{ke2021musiq} and LAION aesthetic score~\cite{laion} with conservative thresholds, yielding targets $x^{\star}$.

\noindent\textbf{Input image generation.} For each target $x^{\star}$, we synthesize a degraded input $x$ by applying random photometric adjustments via a photo-finishing pipeline~\cite{wu2024goal}. This includes perturbations to exposure, gamma, white balance, contrast, tone curves, saturation, shadows/highlights, and HSL. To simulate local retouching, we generate region masks using a Grounding-SAM procedure~\cite{liu2024grounding, ravi2024sam} and additional soft masks from simple priors, applying different degradation parameters within each mask to induce spatially varying edits while preserving fidelity.

\noindent\textbf{Instruction generation.} Given $(x,x^{\star})$, we prompt a multimodal LLM (Qwen2.5-VL-72B~\cite{bai2025qwen2}) in a role-playing template to generate concise, diverse photo-finishing instructions $c_T$ that describe the transformation from $x$ to $x^{\star}$. A small rule-based checker enforces diversity and filters content-edit verbs. Further details on dataset construction are in the Appendix.

\vspace{-5px}
\subsection{Pretrained Multi-step Diffusion}
\label{sec:method-teacher}
\vspace{-5px}

Following InstructPix2Pix~\cite{brooks2023instructpix2pix}, our teacher model $\epsilon_\phi$ is a UNet trained to predict the noise added to a target image's latent representation. Let $x$ be the input image, $x^{\star}$ the target, and $c_T$ the text instruction. We operate in the VAE latent space of a pre-trained Stable Diffusion model~\cite{rombach2022high}, with encoder $\mathcal{E}_\phi$ and decoder $\mathcal{D}_\phi$. During training, noise $\epsilon$ is added to the target latent $z_0 = \mathcal{E}_\phi(x^\star)$ to create a noisy latent $z_t = \alpha_t z_0 + \beta_t \epsilon$. The teacher $\epsilon_\phi$ is trained with an MSE loss to predict this noise, conditioned on the input image latent $c_I = \mathcal{E}_\phi(x)$ and the text prompt $c_T$:
\vspace{-7px}
\begin{equation}
\mathcal{L}_{\text{teacher}}(\phi)
=
\mathbb{E}_{x,x^\star,c_T,t,\epsilon}\!\left[
\big\|
\epsilon - \epsilon_\phi\!\big(z_t,\; t,\; c_I,\; c_T\big)
\big\|_2^2
\right].
\label{eq:teacher-loss}
\vspace{-4px}
\end{equation}
However, applying this multi-step diffusion model for retouching is slow and prone to content drift. We therefore distill from this heavy pretrained editor into a lightweight, one-step retouching model designed to guarantee content fidelity, discussed below.

\vspace{-5px}
\subsection{One-step Bilateral Generator}
\label{sec:method-generator}
\vspace{-5px}

As shown in Fig.~\ref{fig:framework}, our lightweight one-step bilateral grid generator is composed of two branches: a low-resolution diffusion branch and a full-resolution bilateral processing branch. The low-resolution branch contains a frozen VAE encoder $\mathcal{E}_\theta$ and a one-step U-Net denoiser $\epsilon_\theta$, tasked with semantic understanding and preserving the rich diffusion priors. During training, a VAE decoder is temporarily employed to generate a low-resolution image, which helps to stabilize the distillation process. 

The full-resolution branch contains a lightweight bilateral adapter. In a single forward step, it generates a bilateral grid~\cite{chen2007real} $\Gamma\in\mathbb{R}^{H_g\times W_g\times D\times 12}$ that stores local affine transformation parameters in 3D space. This grid is then processed by a fully differentiable ``slice-and-apply" operator that acts on the full-resolution input image of size $(H,W)$. For each input pixel with coordinates $(x', y')$ and color $(r, g, b)$, the operator first computes a grayscale guidance value $z = g(r, g, b)$ via a learned lookup table. It then uses the pixel's spatial coordinates and this guidance value to retrieve a specific affine matrix $A$ by slicing the grid with trilinear interpolation: $A = \Gamma(x'W_g/W, y'H_g/H, z/d)$. Finally, this matrix is applied to the original pixel color, $O = A \cdot (r, g, b, 1)^T$, to produce the final output. This entire mechanism delivers efficient and high-fidelity retouching directly on the high-resolution image.

Our model is highly efficient due to its design. The full-resolution operators have negligible runtime, even at 4K resolution. And the low-resolution branch has constant latency at different resolutions. This enables 4K image processing in just 68ms, vastly outperforming diffusion methods requiring over 10s for 720p inputs.

\vspace{-5px}
\subsection{One-step Bilateral Distillation}
\label{sec:method-onestep}
\vspace{-5px}
Although the proposed one-step generator $G_\theta$ is super-efficient and guarantees no content drift by design, it has a very different structure compared with the pretrained teacher network $\epsilon_\phi$ (diffusion model), posing a challenge in distillation. Therefore, we proposed a novel progressive distillation strategy, described below.

\vspace{-5px}
\subsubsection{Variational Score Distillation in Latent Space}
\label{sec:method-vsd}
\vspace{-5px}

In the low-res one-step diffusion branch, the frozen VAE encoder and decoder are initialized from the weights of pretrained VAE $\mathcal{E}_\phi$ and $\mathcal{D}_\phi$, and the denoising network $\epsilon_\theta$ is initialized from the weights of pretrained denoiser $\epsilon_\phi$.
Recall that diffusion models utilize a UNet to predict the noise $\hat{\epsilon}$ in noisy latent $z_t$, and the denoised latent can be obtained as $\hat{z}_0 = \frac{z_t - \beta_t \hat{\epsilon}}{\alpha_t}$. We directly conducting one-step denoising on the white noise $z_{t_{max}}\sim\mathcal{N}(0,I)$, conditioned on $c_I=\mathcal{E}_\theta(x)$ and $c_T$, to predict the clean latent $\hat{z}_0$ is calculated as:
\vspace{-12px}
\begin{equation}
\hat{z}_0 = \frac{z_{t_{max}} - \beta_t \epsilon_\theta(z_{t_{max}},\; t_{max},\; c_I,\; c_T)}{\alpha_t}, 
\label{eq:onestep-pred}
\vspace{-6px}
\end{equation}
and the corresponding low-resolution image is $\hat{x}=\mathcal{D}_\theta(\hat{z}_0)$. Note that during inference, the VAE decoder is not used, as $\hat{x}$ only helps to stabilize the distillation process, and is not needed in full-res bilateral processing. We regularize $\epsilon_\theta$ with a latent-space Variational Score Distillation (VSD) loss, $\mathcal{L}_{\text{VSD}}$, following the design in DMD~\cite{yin2024dmd}.

VSD loss introduces a trainable regularizer $\epsilon_{\phi'}$ finetuned on the distribution of generated images $\hat{x}$ of the one-step generator $\epsilon_{\theta}$ to replicate its behaviour. Given the clean latent predicted by the one-step generator via Eqn.~\ref{eq:onestep-pred}, we add noise to it to construct the noisy latent $\hat{z}_t=\alpha_t \hat{z}_0 + \beta_t \epsilon$. This $\hat{z}_t$ then serves as a common input to the teacher and regularizer to compute a stable gradient that steers the student towards the teacher.
We adopt the latent form of VSD used in DMD~\cite{yin2024dmd, wu2024one}. The gradient of the VSD loss w.r.t.\ $\theta$  $\nabla_\theta \mathcal{L}_{\text{VSD}}$ is 
\vspace{-10px}
\begin{equation}
\begin{aligned}
\mathbb{E}_{t,\epsilon,\hat{z}_t}\!\Big[
\omega(t)\big(
\epsilon_\phi(\hat{z}_t, t, c_I, c_T) -
\epsilon_{\phi'}(\hat{z}_t, t, c_I, c_T)
\big)
\frac{\partial \hat{z}_0}{\partial \theta}
\Big].
\end{aligned}
\label{eq:vsd-two-cond}
\end{equation}

To ensure the regularizer $\epsilon_{\phi'}$ remains a faithful proxy for the generator's current state, it is trained concurrently on noisy samples $\hat{z}_t$ derived from the generator's own outputs $\hat{z}_0$:
\vspace{-12px}
\begin{equation}
\mathcal{L}_{\text{diff}}(\phi')
=
\mathbb{E}_{t,\epsilon,c_I,c_T,\hat{z}_t}\!\left[
\left\|
\epsilon_{\phi'}\!\left(\hat{z}_t, t, c_I, c_T\right)
-
\epsilon
\right\|_2^2
\right].
\label{eq:reg-two-cond}
\end{equation}
\vspace{-20px}

To further stabilize this process, we adopt a progressive schedule. Training begins with high noise levels ($t\!\in\![t_{\text{hi}},t_{max}]$) to learn coarse attributes like tone and exposure, before we gradually lower $t_{\text{hi}}$ to distill fine-grained color details.

\subsubsection{Prompt Alignment Loss}
\label{sec:method-clip}

Distilling a multi-step editor into one step often weaken the coupling between the instruction $c_T$ and often yields ``plausible but misdirected'' retouches under weak, aesthetic instructions. Thus, we need to add further supervision to ensure the output image follows the users' instructions.

Specifically, unlike object edits, retouching intents are mostly \emph{directional and compositional} (e.g., warmer, dreamy, cinematic). While the VSD loss and data loss ensure feasibility, they do not guarantee that the change follows the intended semantic direction. We therefore convert user instruction $c_T$ into a small set of atomic \emph{retouching attributes} $\mathcal{A}(c_T)=\{a\}$ using a rule-based matcher. Each attribute is an explicit edit direction tailored to photo retouching (e.g., \texttt{brightness:up}, \texttt{contrast:down}, \texttt{mood:cozy}, \texttt{temperature:warm}, \texttt{style:vintage}) and is paired with two short text prompts describing positive and negative directions. (e.g., ``Bright Image'' vs.\ ``Dark Image''). This \emph{attribute bank} turns a long, weak instruction into several stable, additive supervision signals. Let $\mathbf{e}^{\text{img}}(\cdot)$ and $\mathbf{e}^{\text{text}}(\cdot)$ be frozen CLIP image and text encoders. The cosine similarity of the two $\ell_2$-normalized image and text embeddings is used and its scalar value is denoted by $s$. For an attribute $a$ (e.g., \texttt{mood:cozy}) with prompts $(p_a^+,p_a^-)$
, define $s_a^+=\langle \mathbf{e}^{\text{img}}(\hat{x}),\,\mathbf{e}^{\text{text}}(p_a^+)\rangle$ and $s_a^-=\langle \mathbf{e}^{\text{img}}(\hat{x}),\,\mathbf{e}^{\text{text}}(p_a^-) \rangle$, where $\langle \cdot,\cdot\rangle$ denotes cosine similarity. The per-attribute InfoNCE loss~\cite{oord2018representation} (viewed as a function of $a$) is
\vspace{-5px}
\begin{equation}
\ell_{\text{nce}}(a)
= -\log\frac{\exp\!\big(s_a^+/\tau\big)}
{\exp\!\big(s_a^+/\tau\big)+\exp\!\big(s_a^-/\tau\big)}.
\label{eq:attr-infonce}
\vspace{-5px}
\end{equation}
Finally, with confidences $w_a$ from the matcher, the language alignment loss is applied to the one-step branch during distillation:
\vspace{-12px}
\begin{equation}
\mathcal{L}_{\text{align}}
=\frac{1}{|\mathcal{A}(c_T)|}\sum_{a\in\mathcal{A}(c_T)}
\Big[w_a\,\ell_{\text{nce}}(a)
\Big].
\label{eq:attr-total}
\vspace{-6px}
\end{equation}
This supervision restores directional alignment lost by step compression,  and resolves ambiguity among many color transforms that could otherwise minimize the data term and VSD term yet deviate from $c_T$.

\subsubsection{Data Supervision Loss}
\label{sec:method-data}
To stabilize distillation, we also add a data term that supervises the low-resolution output $\hat{x}$ with the ground truth target $x^{\star}$, following~\cite{yin2024dmd, wu2024one}:
\vspace{-6px}
\begin{equation}
\mathcal{L}_{\text{data}}=
\|\hat{x}-x^{\star}\|_2^2
+\lambda_{\text{LPIPS}}\mathcal{L}_{\text{LPIPS}}(\hat{x},x^{\star}).
\label{eq:data}
\vspace{-4px}
\end{equation}

\subsubsection{Bilateral Loss}
\vspace{-3pt}
\label{sec:method-bila}
The losses above focus on training the low-resolution branch. To also train the full-resolution bilateral processing branch, we introduce $\mathcal{L}_{\text{bila}}$. Let $\hat{x}_B$ be the final high-resolution output. This loss includes: (i) $\ell_1$ and LPIPS losses against the ground truth $x^\star$, (ii) a perceptual agreement term encouraging $\hat{x}_B$ to match the low-res prediction $\hat{x}$, and (iii) a laplacian regularizers on the bilateral grid $\Gamma$ for smoothness and a penalty that prevents RGB overflow:
\vspace{-6px}
\begin{align}
\mathcal{L}_{\text{bila}}&=\lambda_{1}\|\,\hat{x}_B-x^\star\,\|_1+\lambda_{2}\cdot\mathcal{L}_\text{LPIPS}(\hat{x}_B,x^\star) \nonumber\\
&\quad + \lambda_{3}\cdot\mathcal{L}_\text{LPIPS}\big(\hat{x}_B,\hat x\big) \nonumber\\
&\quad + \lambda_{4}\cdot\|\Delta^3 {\Gamma}\|_2^2 + \lambda_{5}\cdot\Psi(\hat{x}_B),
\vspace{-16px}
\end{align}
where $\Delta^3$ is a 3D Laplacian regularizer to penalize differences between adjacent cells over the bilateral grid for smoothness, and $\Psi$ is a soft penalty discouraging out-of-gamut RGB.

\subsubsection{Overall Objective and Distillation Strategy}
\label{sec:method-all-loss}

Combining all training losses above, we finally design a novel two-stage progressive distillation strategy.

\noindent\textbf{Stage 1: Low-Resolution one-step diffusion branch training.} In this stage, we only train the low-resolution one-step diffusion branch. Note that the low-resolution branch shares the same network structure as the pretrained diffusion and thus distillation training is easier compared with the bilateral processing network. During training, we optimize the U-Net $\epsilon_\theta$ and the VSD regularizer $\epsilon_{\phi'}$. The objective combines VSD, the data term, and our prompt alignment loss:
\vspace{-10px}
\begin{equation}
\mathcal{L}_{\text{stage1}}= \mathcal{L}_{\text{data}}+\lambda_{\text{VSD}}\mathcal{L}_{\text{VSD}}+\lambda_{\text{align}}\mathcal{L}_{\text{align}}.
\vspace{-5px}
\end{equation}

\noindent\textbf{Stage 2: Joint bilateral distillation.} After the first stage converges, we unfreeze the bilateral adapter and train the entire generator end-to-end. Since stage 1 already trains the relative heavy low-resolution network, finetuning the lightweight full-resolution bilateral processing is also trackable. To train the bilateral processing, a bilateral loss is added:
\vspace{-4px}
\begin{equation}
\mathcal{L}_{\text{stage2}}=\mathcal{L}_{\text{stage1}}+\lambda_{\text{bila}}\mathcal{L}_{\text{bila}}.
\vspace{-5px}
\end{equation}
This complete \emph{one-step bilateral distillation} framework yields an efficient model that guarantees high-fidelity, content-preserving retouching while retaining strong instruction-following capabilities.

\vspace{-8px}
\section{Experiments}
\vspace{-5px}
\subsection{Experiment Setup}
\vspace{-5px}

\begin{table*}[!ht]
\centering
\small
\vspace{-10px}
\caption{Comparison on iRetouch benchmark. Our method achieves state-of-the-art efficiency and content fidelity while remaining highly competitive in editing quality. Blank entries indicate models that cannot process high resolutions or are not instruction-driven.}
\vspace{-6px}

\setlength{\tabcolsep}{5pt}
\begin{adjustbox}{width=\textwidth}
\begin{tabular}{c|cccc cccc ccccc}
\toprule
\multirow{2}{*}{Method} & \multicolumn{4}{c}{Runtime(s)} & \multicolumn{4}{c}{Content Fidelity} & \multicolumn{5}{c}{Editing Quality} \\
\cmidrule(lr){2-5}\cmidrule(lr){6-9}\cmidrule(lr){10-14}
& 720p$\downarrow$ & 1K$\downarrow$ & 2K$\downarrow$ & 4K$\downarrow$ & SSIM$\uparrow$ & CW-SSIM$\uparrow$ & GSMD$\downarrow$ & DISTS$\downarrow$ & L1$\downarrow$ & L2$\downarrow$ & SC$\uparrow$ & PQ$\uparrow$ & O$\uparrow$ \\
\midrule
3DLUT~\cite{zeng20203dlut}             & 0.066 & 0.079 & 0.112 & 0.201 & 0.982 & \textbf{0.981} & 0.013 & 0.024 & 0.136  & 0.034   & -  & -    & - \\
RSFNet~\cite{ouyang2023rsfnet}         & \textbf{0.029} & \textbf{0.047} & 0.086 & 0.189 & 0.975 & 0.976 & \textbf{0.012} & 0.038 & 0.137  & 0.034    & -  & -    & - \\
\midrule
InstructPix2Pix~\cite{brooks2023instructpix2pix}  & 4.632 & - & - & - & 0.742 & 0.768 & 0.149 & 0.177  & 0.164 & 0.050 & 7.11 & 7.58 & 7.34 \\
Step1X\mbox{-}Edit~\cite{liu2025step1x}          & 57.932 & - & - & - & 0.706 & 0.694 & 0.174 & 0.167  & 0.140 & 0.036 & 7.63 & 8.52 & 8.06 \\
GPT\mbox{-}Image\mbox{-}1~\cite{hurst2024gpt4o}  & 15.427 & 21.889 & - & - & 0.505 & 0.397 & 0.242 & 0.216  & 0.215 & 0.082 & 8.09 & 8.56 & 8.32 \\
Qwen\mbox{-}Image~\cite{wu2025qwen}              & 7.720 & - & - & - & 0.689 & 0.744 & 0.174 & 0.147  & 0.168 & 0.054 & 8.12 & 8.67 & 8.39 \\
FLUX.1\mbox{-}Kontext\mbox{-}Pro~\cite{labs2025flux} & 10.235 & - & - & - & 0.802 & 0.857 & 0.112 & 0.132  & 0.161 & 0.050 & 7.56 & 8.72 & 8.12 \\
Gemini\mbox{-}2.5\mbox{-}Flash~\cite{comanici2025gemini} & 14.440 & - & - & - & 0.676 & 0.796 & 0.175  & 0.115 & 0.137 & 0.036 & \textbf{8.56} & 8.94 & \textbf{8.74} \\
\midrule
\textbf{Ours}              & 0.065 & 0.065 & \textbf{0.066} & \textbf{0.068} & \textbf{0.989} & 0.973 & \textbf{0.012} & \textbf{0.022}  & \textbf{0.099} & \textbf{0.018} & 8.14 & \textbf{8.98} & 8.54 \\
\bottomrule
\end{tabular}
\end{adjustbox}
\label{tab:main-res}
\end{table*}

\noindent\textbf{New benchmark iRetouch.}
For evaluation, we have created a new benchmark, iRetouch, consisting of 500 real-world before-and-after retouching pairs from the Adobe Lightroom community. Instructions for these pairs are generated using our method from~\cref{sec:finetune}, followed by manual refinement for clarity and diversity.
The benchmark spans a wide variety of scenes (e.g., portraits, landscapes) and includes a rich vocabulary of retouching edits, such as global adjustments, specific styles (cinematic, dreamy), moods, and local effects (see Appendix for a detailed breakdown).

\noindent\textbf{Content fidelity metrics.} Retouching is non-destructive, so edits must preserve structure and texture without repaints. To factor out intentional tone changes, we convert outputs to grayscale and histogram-match them to the input, then compute SSIM~\cite{wang2004ssim} (structural similarity), CW-SSIM~\cite{sampat2009complexwssim} (geometry and texture distortion), DISTS~\cite{ding2020dists} (textural similarity), and GMSD~\cite{xue2013gmsd} (gradient-magnitude consistency).

\noindent\textbf{Editing quality metrics.} Following prior work~\cite{liu2025step1x, zhang2023magicbrush}, we report L1/L2 distances, instruction–image alignment (SC, 0–10), perceptual quality (PQ, 0–10), and the overall score O = $\sqrt{\text{SC}\times\text{PQ}}$. SC and PQ are generated using GPT-4o, similar to~\cite{liu2025step1x}. Additional details are provided in the Appendix.

\noindent\textbf{Implementation.} Our one-step bilateral generator is built upon a pre-trained Stable Diffusion editor. We initialize our student U-Net from the teacher's weights and freeze the VAE. VSD distillation follows a three-stage curriculum over timesteps to learn from coarse structure and tone (high $t$), then instruction alignment (mid $t$), and finally fine-grained color details (low $t$).
Training is at 512px using AdamW~\cite{loshchilov2017decoupled} with EMA, mixed precision, and gradient clipping. Inference is a single pass: the model generates a bilateral grid and applies it to the native resolution input, yielding constant-time performance regardless of image size. We train on our instruction-retouching dataset in Sec.~\ref{sec:finetune}.
See the Appendix for full implementation details.

\noindent\textbf{Runtime.} We measure end-to-end latency across resolutions from 720p to 4K. Open-source models are benchmarked on a server with 8 NVIDIA RTX 4090 GPUs. For proprietary models, we report the full end-to-end API latency, including data transfer.

\vspace{-6px}
\subsection{Evaluation and Results}
\vspace{-6px}
We compare our method with baselines across three categories: (1) traditional enhancement methods~\cite{zeng20203dlut, ouyang2023rsfnet}, (2) open-source image editing models~\cite{brooks2023instructpix2pix, liu2025step1x, wu2025qwen}, and (3) proprietary large-scale editing models~\cite{hurst2024gpt4o, labs2025flux, comanici2025gemini}.

\vspace{-4px}
\subsubsection{Evaluation on Our iRetouch Benchmark}
\vspace{-5px}
As shown in Tab.~\ref{tab:main-res}, our method outperforms others in terms of runtime, fidelity, and editing quality.

\noindent\textbf{Efficiency.} Our model demonstrates exceptional efficiency, maintaining a near-constant inference time of 0.065–0.068s from 720p up to 4K resolutions. This represents a 70–900$\times$ speedup over generative baselines at 720p. The blank runtime entries for some baselines highlight a critical limitation: most diffusion-based models cannot natively process resolutions beyond 1K, a barrier our design overcomes.

\noindent\textbf{Fidelity.} Our approach achieves state-of-the-art content fidelity among all instruction-guided models. This confirms our bilateral branch successfully prevents the textural distortions common in pure diffusion editors. 

\noindent\textbf{Quality.} For editing quality, our model's overall score (O) of 8.54 is highly competitive with the top proprietary system (Gemini-2.5-Flash at 8.74) and significantly surpasses other open-source editors. The blank quality scores for traditional methods like 3D LUT exist because they are not instruction-driven and thus cannot be evaluated for semantic alignment. In summary, our method delivers near–state-of-the-art editing quality with state-of-the-art fidelity and 4K-constant runtime. These results support our design goal: instruction-guided retouching that is high-fidelity, fast, and stable across resolutions.
We also provide a more detailed analysis of the relationship between quality and fidelity in the Appendix.

\noindent\textbf{Visual comparison.}
\cref{fig:exp-main} provides a qualitative comparison across a range of instructions. The results highlight a common failure in competing methods: a trade-off between editing quality and content fidelity. Generative editors like InstructPix2Pix and GPT-Image-1 often introduce severe artifacts, hallucinations, or unwanted text overlays, fundamentally altering the source image. Even capable models like Gemini-2.5-Flash can subtly change key features. Our method, however, successfully follows both global and local instructions while maintaining high fidelity, applying the desired stylistic edits without distorting content or compromising the original photograph's integrity.

\begin{figure*}[!ht]
\centering
  \includegraphics[width=1.0\textwidth]{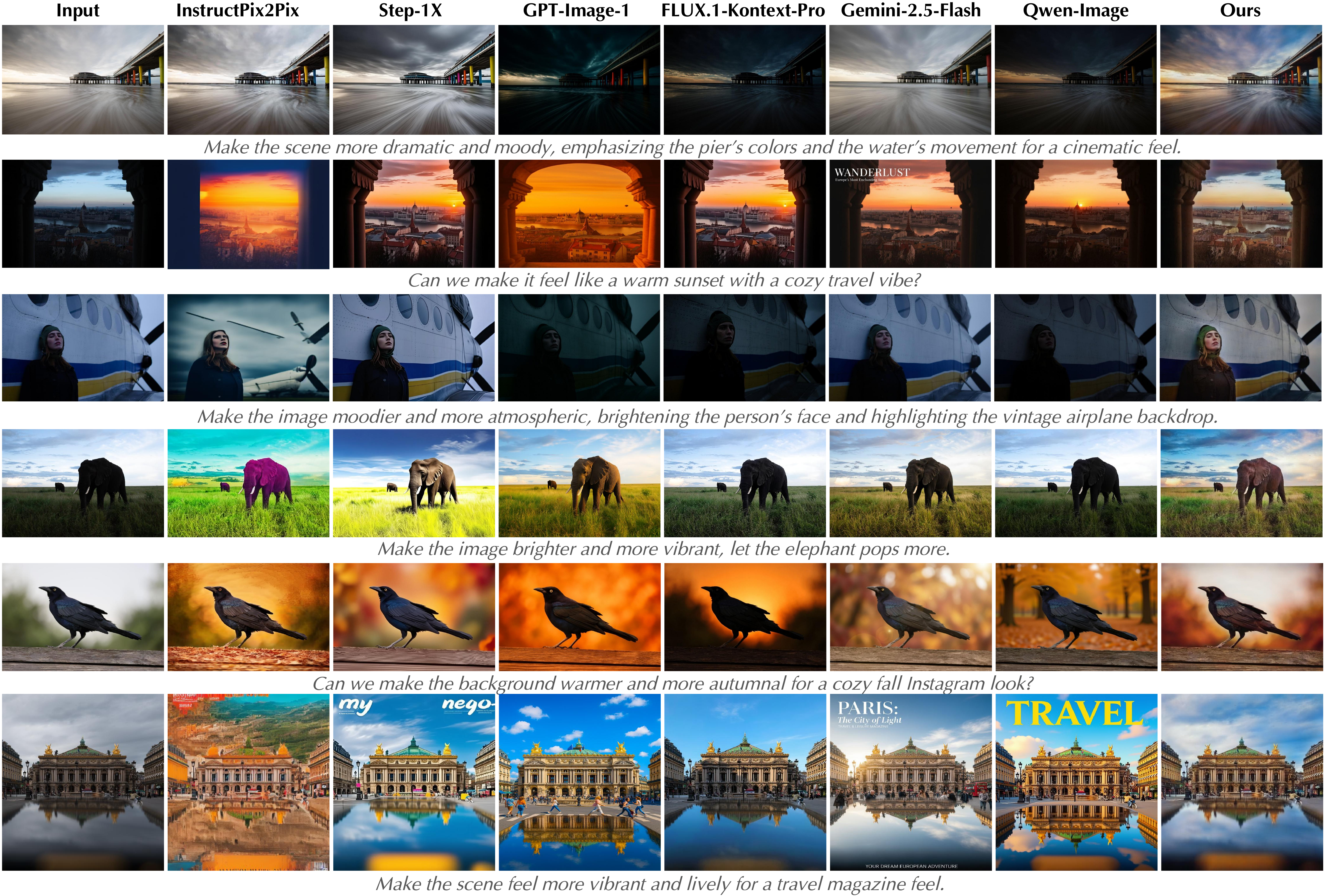}
  \vspace{-23pt}
  \caption{Visual comparisons of different image editing methods on our iRetouch benchmark.}
  \vspace{-15pt}
  \label{fig:exp-main}
\end{figure*}

\vspace{-4px}
\subsubsection{User Study}
\vspace{-4px}

\begin{figure}[ht]
\centering
  \includegraphics[width=0.95\linewidth]{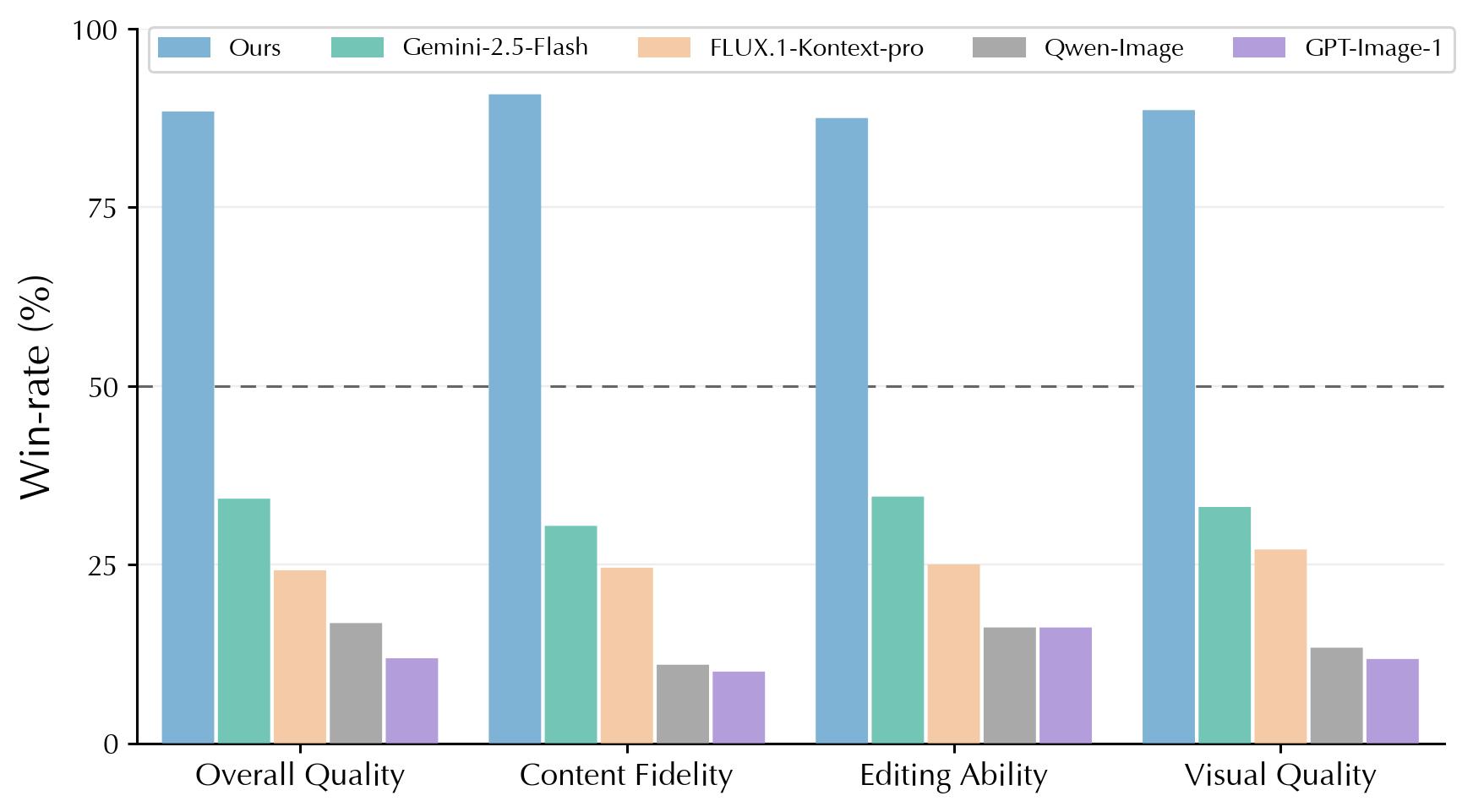}
  \vspace{-10pt}
  \caption{User preference study results on iRetouch benchmark.}
  \vspace{-8pt}
  \label{fig:user-study}
\end{figure}

To assess subjective user preference, we conducted a user study with 30 participants, who evaluated 20 retouching examples from our iRetouch benchmark. They compared our method against four leading baselines (FLUX.1-Kontext-pro, Gemini-2.5-Flash, Qwen-Image, GPT-Image-1) across four dimensions: content fidelity, editing ability, visual quality, and overall preference. As shown in Fig.~\ref{fig:user-study}, the results reveal a clear and consistent preference for our method. Our approach achieved the highest ratings in all categories, confirming that users favor its artifact-free, high-fidelity edits that accurately reflect their intent.

\subsubsection{Evaluation of Identity Preservation on PPR10K}

\begin{figure}[ht]
\centering
  \includegraphics[width=0.99\linewidth]{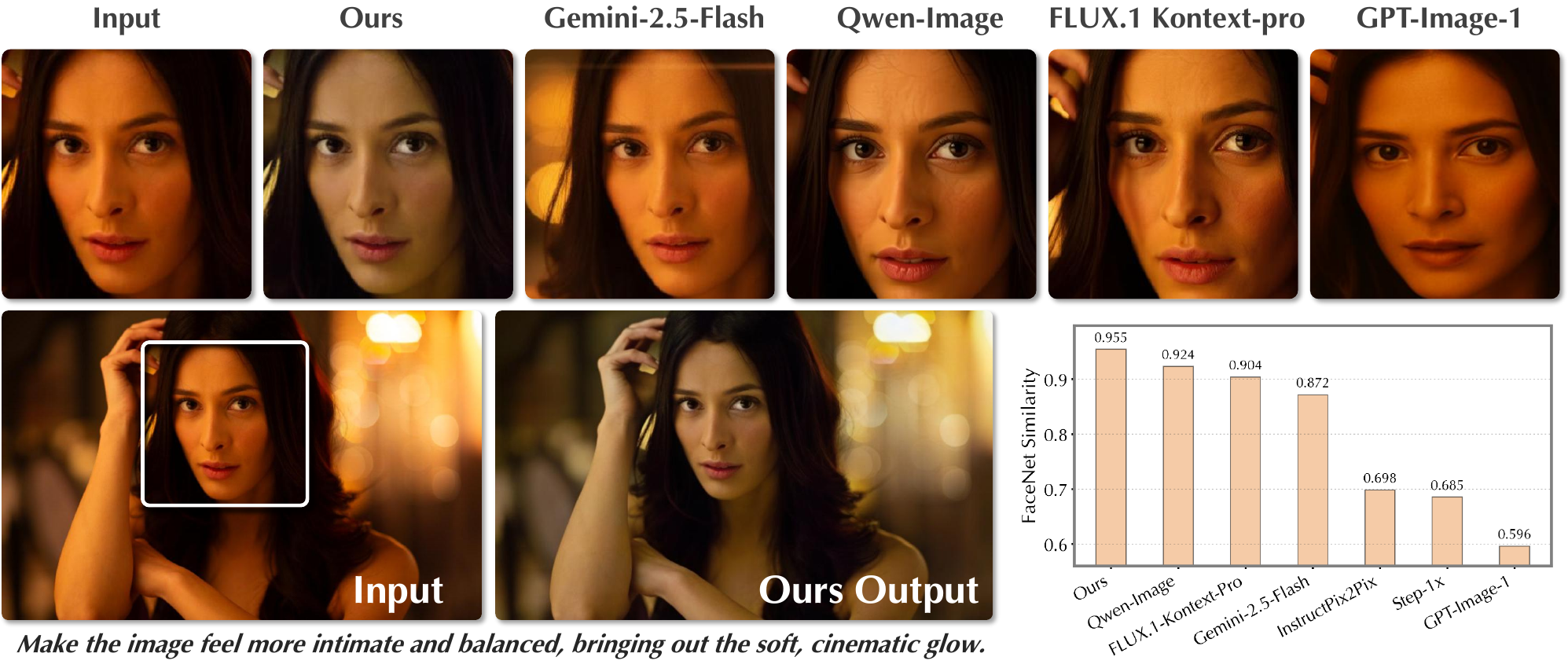}
  \vspace{-5pt}
  \caption{Results of identity preservation comparison on the PPR10K dataset. Our model scores highest in facial similarity and avoids the identity-altering artifacts.}
  \vspace{-15pt}
  \label{fig:exp-face-vis}
\end{figure}

In portrait editing, content fidelity is crucial as it requires strict identity preservation. To evaluate identity preservation on this task, we test on 100 images from the PPR10K dataset~\cite{liang2021ppr10k} with MLLM generated instructions. 
We quantify identity preservation by extracting facial embeddings from the input and output images using FaceNet~\cite{schroff2015facenet} and then computing their cosine similarity. As shown quantitatively in Fig.~\ref{fig:exp-face-vis}, our method achieves the highest face similarity score. We also include qualitative comparison in the figure; our model retouches the portrait while strictly preserving fidelity, whereas competing methods introduce noticeable repainting that distorts the subject's identity.

\vspace{-5pt}
\subsection{Ablation}
\vspace{-5pt}

\begin{table}[!ht]
\centering
\vspace{-5px}
\caption{Ablation study on our framework. We evaluate content fidelity, editing quality, and runtime. Our full model effectively combines the strengths of diffusion priors and bilateral processing, achieving high scores across all criteria.}
\vspace{-10px}
\label{tab:ablation_single}
\begin{adjustbox}{width=\linewidth}
\begin{tabular}{l|c ccc ccc}
\toprule
\multirow{2}{*}{Method} & \multirow{2}{*}{Runtime(s)$\downarrow$} & \multicolumn{3}{c}{Content Fidelity} & \multicolumn{3}{c}{Editing Quality} \\
\cmidrule(lr){3-5}\cmidrule(lr){6-8}
& & SSIM$\uparrow$ & GSMD$\downarrow$ & DISTS$\downarrow$ & SC$\uparrow$ & PQ$\uparrow$ & O$\uparrow$ \\
\midrule
Bilateral Grid Prediction & 0.001 & 0.996 & 0.003 & 0.005 & 4.48 & 8.28 & 6.09 \\
Teacher (Multi-step Diffusion) & 4.602 & 0.833 & 0.095 & 0.121 & 7.96 & 8.71 & 8.33 \\
Hybrid (Teacher Features + Bilateral) & 0.065 & 0.904 & 0.073 & 0.107 & 5.65 & 7.62 & 6.56 \\
Student (Diffusion-Only) & 0.319 & 0.788 & 0.130 & 0.152 & 8.43 & 8.85 & 8.64 \\
\textbf{Ours (Full Model)} & \textbf{0.065} & \textbf{0.989} & \textbf{0.012} & \textbf{0.022} & \textbf{8.14} & \textbf{8.98}  & \textbf{8.54} \\
\bottomrule
\end{tabular}
\end{adjustbox}
\vspace{-10px}
\end{table}

We conduct a series of ablation studies to validate our key design choices, focusing on our framework and the components of our distillation algorithms.

\noindent\textbf{Ablation on framework.} We first analyze the contribution of our framework components in Tab.~\ref{tab:ablation_single}. We compare our full model against four key baselines: (1) \emph{Bilateral Grid Prediction}, a model that directly predicts a bilateral grid from the input image without diffusion priors, trained on our dataset; (2) our \emph{Teacher (Multi-step Diffusion)} model; (3) a \emph{Hybrid} model that uses features from the multi-step teacher to predict a bilateral grid; and (4) our \emph{Student (Diffusion-Only)}, which corresponds to the low-resolution RGB output from our distilled U-Net without the bilateral branch.

The results in Tab.~\ref{tab:ablation_single} reveal a clear trade-off. Purely diffusion-based models (Teacher, Student-Only) achieve high editing quality but low fidelity. In contrast, a simple Bilateral Grid Prediction model preserves content perfectly (0.996 SSIM) but fails to follow instructions (6.09 O-score). Our full model uniquely resolves this conflict by merging the semantic strength of diffusion (8.54 O-score) with the structural preservation of bilateral processing (0.989 SSIM), all while maintaining high efficiency. This validates our dual-branch design for balancing fidelity, quality, and speed.

\begin{figure}[t]
\centering
  \includegraphics[width=1.0\linewidth]{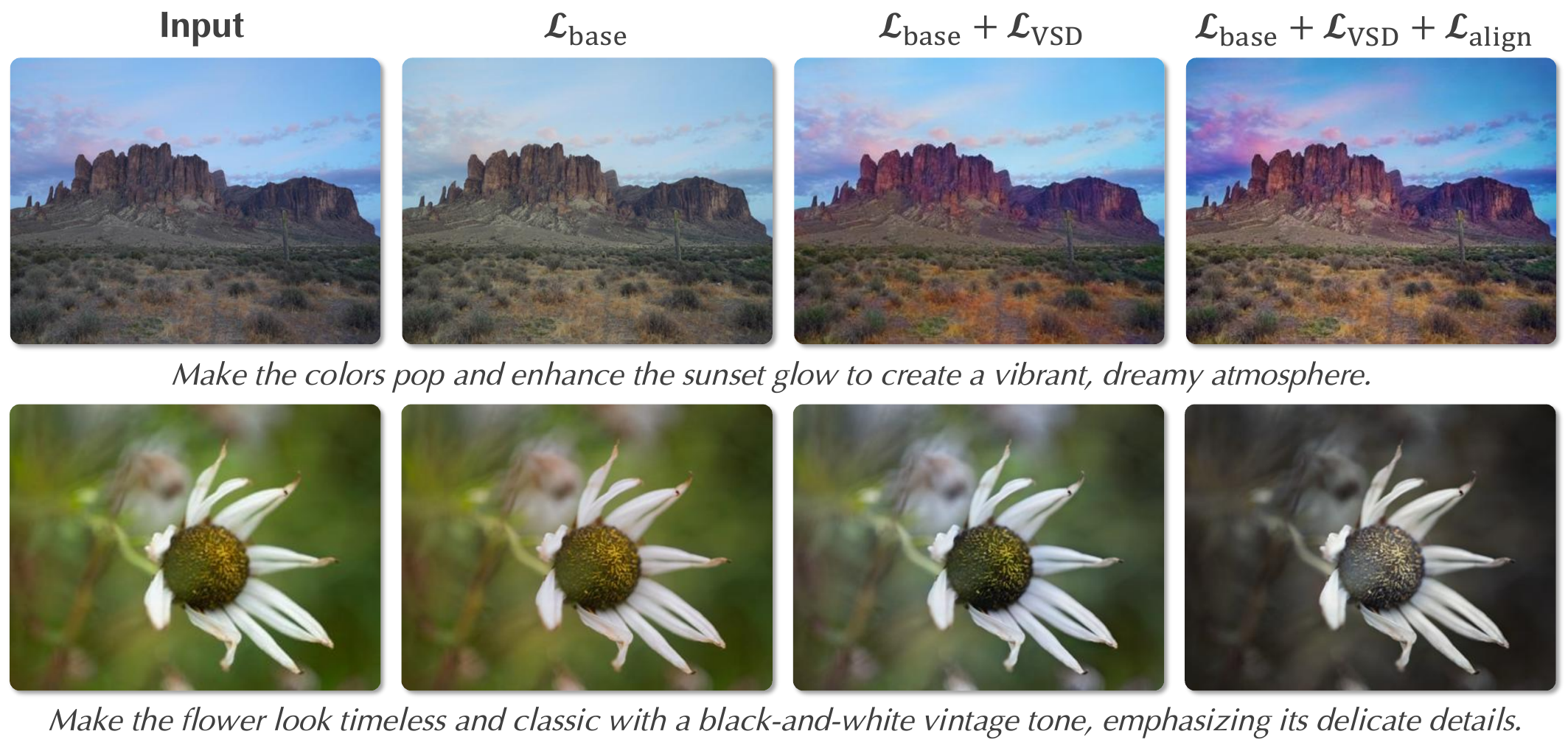}
  \vspace{-22pt}
  \caption{Visualization of ablation study on the loss configuration of one-step bilateral distillation.}
  \vspace{-12pt}
  \label{fig:exp-abla-vis}
\end{figure}

\begin{table}[t]
\centering
\small
\setlength{\tabcolsep}{6pt}
\caption{Ablation on our distillation loss components. Both VSD and our prompt alignment loss ($\mathcal{L}_{\mathrm{align}}$) are critical for achieving high editing quality.}
\vspace{-10px}
\label{tab:ablation_loss_components}
\begin{adjustbox}{width=0.7\linewidth}
\begin{tabular}{c|ccc}
\toprule
\multirow{2}{*}{Loss Configuration} & \multicolumn{3}{c}{Editing Quality} \\
\cmidrule(lr){2-4}
& SC$\uparrow$ & PQ$\uparrow$ & O$\uparrow$ \\
\midrule
$\mathcal{L}_{\mathrm{base}}$ & 5.978 & 8.280 & 7.036 \\
$\mathcal{L}_{\mathrm{base}} + \mathcal{L}_{\mathrm{VSD}}$ & 7.257 & 9.013 & 8.087 \\
$\mathcal{L}_{\mathrm{base}} + \mathcal{L}_{\mathrm{VSD}} + \mathcal{L}_{\mathrm{align}}$ &  8.140 & 8.984 & 8.553 \\
\bottomrule
\end{tabular}
\end{adjustbox}
\vspace{-10px}
\end{table}

\noindent\textbf{Ablation on one-step bilateral distillation.} 
Next, we validate the effectiveness of the core loss components in our one-step bilateral distillation process. As shown in~\cref{tab:ablation_loss_components}, we start with a base objective, $\mathcal{L}_{\mathrm{base}}$, which includes only the data term and bilateral losses ($\mathcal{L}_{\mathrm{data}} + \mathcal{L}_{\mathrm{bila}}$). We then progressively add our main distillation loss, $\mathcal{L}_{\mathrm{VSD}}$, and our prompt alignment loss, $\mathcal{L}_{\mathrm{align}}$.

As shown in Tab.~\ref{tab:ablation_loss_components}, the base model ($\mathcal{L}_{\mathrm{base}}$) alone is insufficient for quality editing. Adding $\mathcal{L}_{\mathrm{VSD}}$ is critical, dramatically boosting the score by transferring the teacher's generative priors. Incorporating our prompt alignment loss ($\mathcal{L}_{\mathrm{align}}$) provides a final, significant gain. This confirms its role in providing essential directional supervision for interpreting stylistic prompts where VSD alone falls short. We also visualize this ablation in Fig.~\ref{fig:exp-abla-vis}.

\subsection{Fine-grained Control}
\vspace{-10pt}

\begin{figure}[!ht]
\centering
  \includegraphics[width=0.99\linewidth]{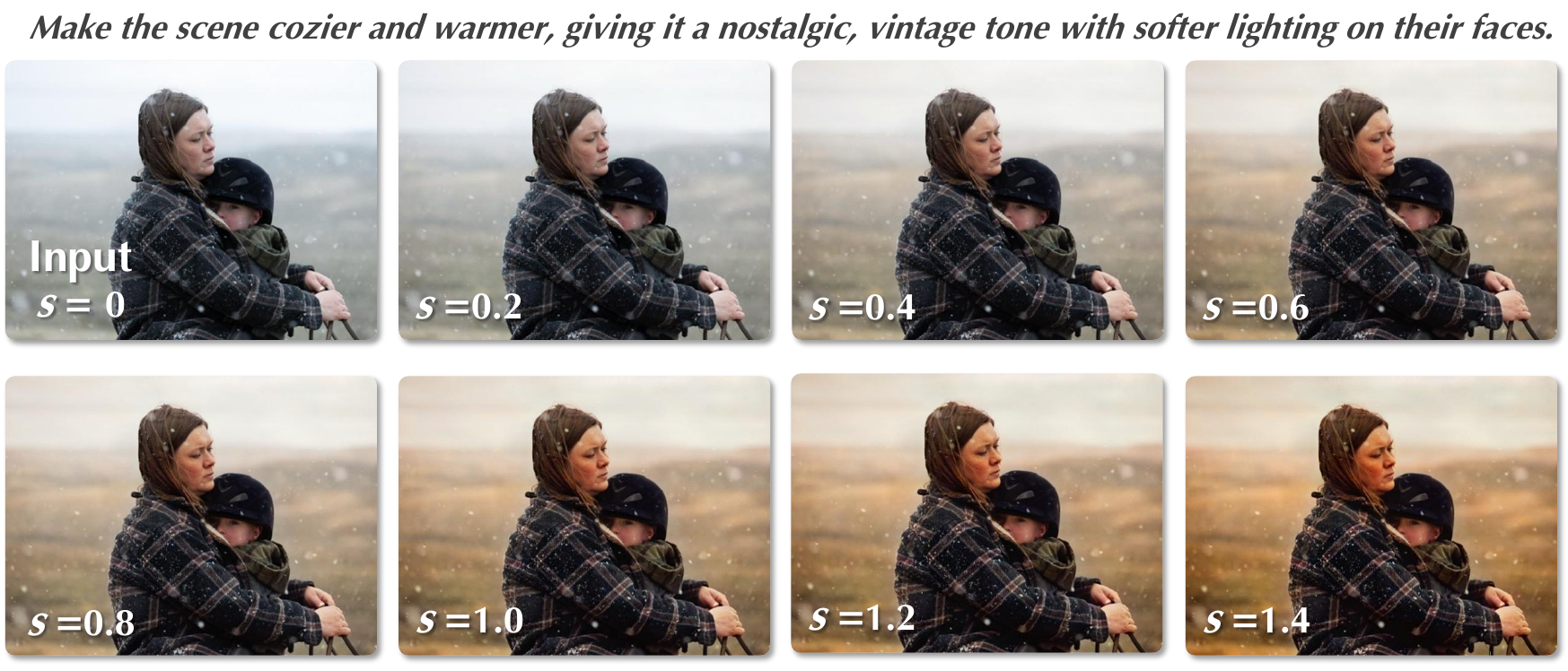}
  \vspace{-10pt}
  \caption{Visualization of continuous control on editing strength.}
  \vspace{-12pt}
  \label{fig:exp-ext-1}
\end{figure}

Our framework's control extends beyond language prompts to include more fine-grained control over the retouching effect. As shown in Fig.~\ref{fig:exp-ext-1}, users can continuously adjust the retouching intensity by applying a scalar $s$ to the per-pixel affine transforms. Thanks to the linearity of the affine transform in bilateral space, setting $s=0$ yields the input, while $s>1$ enhances the effect. This transforms our model into a smart, language-guided filter, offering precise control where language can be ambiguous. Moreover, we support fine-grained regional control using a soft bilateral blending strategy, which is further detailed in the Appendix.

\section{Conclusion}
In this work, we present an efficient and fidelity-preserving approach to image retouching that addresses both fidelity degradation and computational inefficiency. Instead of manipulating pixels or latent features, our method operates in a compact, content-decoupled bilateral space, enabling high fidelity with significantly improved efficiency.
To preserve strong generative priors, we distill a multi-step diffusion model into our bilateral grid framework via variational score distillation, enhanced with a CLIP-based contrastive loss for instruction following. We further introduce a new benchmark dataset for instruction-guided retouching and evaluate fidelity, instruction alignment, and efficiency.
Compared to recent image editing methods such as Gemini-2.5-Flash (Nano Banana), our approach runs orders of magnitude faster while achieving superior content fidelity and comparable instruction-following performance.

\section{Acknowledgement}
This study was supported in part by the Shanghai Artificial Intelligence Laboratory, the Centre for Perceptual and Interactive Intelligence (CPII) Ltd., a CUHK-led InnoCentre under the InnoHK initiative of the Innovation and Technology Commission of the Hong Kong Special Administrative Region Government. The work is supported by the National Key R\&D Program of China (No. 2025YFE0201300). We thank Xin Cai and Zixuan Chen for helpful discussions.

{
    \small
    \bibliographystyle{ieeenat_fullname}
    \bibliography{main}
}


\end{document}